\documentclass[letterpaper]{article}
\usepackage{flushend}
\usepackage{aaai}
\usepackage{times}
\usepackage{helvet}
\usepackage{courier}
\usepackage{graphicx}

\usepackage{amsmath}
\usepackage{amssymb}
\usepackage{amsfonts}
\usepackage{amsthm}
\usepackage{dsfont}

\usepackage{epsfig}
\usepackage{xspace}

\newtheorem{theorem}{Theorem}
\newtheorem{proposition}[theorem]{Proposition}

\newtheorem{example}{Example}
\newtheorem{axiom}{Axiom}
\newtheorem{property}{P\!}

\newcommand{\beq}{\begin{equation}}
\newcommand{\eeq}{\end{equation}}
\newcommand{\bea}{\begin{eqnarray}}
\newcommand{\eea}{\end{eqnarray}}
\newcommand{\bean}{\begin{eqnarray*}}
\newcommand{\eean}{\end{eqnarray*}}

\newcommand{\argmin}{\operatornamewithlimits{arg\,min}}

\newcommand{\cost}{L}

\newcommand{\partition}{g}
\newcommand{\work}{w}
\newcommand{\ques}{x}

\newcommand{\lowerw}{{W_{\min}}}
\newcommand{\upperw}{{W_{\max}}}
\newcommand{\spammerw}{0}
\newcommand{\workinterval}{[\lowerw,\upperw]}

\newcommand{\fname}{objective\xspace}

\newcommand{\numitems}{d}
\newcommand{\numworkers}{k}
\newcommand{\numobs}{n}

\newcommand{\workbold}{\boldsymbol{\work}}
\newcommand{\quesbold}{\boldsymbol{\ques}}

\newcommand{\respmx}{Y}

\emergencystretch=\maxdimen 

\begin{document}
%
\title{On the Impossibility of Convex Inference in Human Computation}
\author{}
\author{Nihar B. Shah\\U.C. Berkeley \\ nihar@eecs.berkeley.edu \And Dengyong Zhou \\ Microsoft Research \\ dengyong.zhou@microsoft.com}
\maketitle
\thispagestyle{empty}
\pagestyle{empty}
\begin{abstract}
\begin{quote}

Human computation or crowdsourcing involves joint inference of the  ground-truth-answers and the worker-abilities by optimizing an objective function, for instance, by maximizing the data likelihood based on an assumed underlying model. A variety of methods have been proposed in the literature to address this inference problem. As far as we know, none of the objective functions in existing methods is convex. In machine learning and applied statistics, a convex function such as the objective function of support vector machines (SVMs) is generally preferred, since it can leverage the high-performance algorithms and rigorous guarantees established in the extensive literature on convex optimization. One may thus wonder if there exists a meaningful convex objective function for the inference problem in human computation. In this paper, we investigate this convexity issue for human computation. We take an axiomatic approach by formulating a set of axioms that impose two mild and natural assumptions on the objective function for the inference. Under these axioms, we show that it is unfortunately impossible to ensure convexity of the inference problem. On the other hand, we show that interestingly, in the absence of a requirement to model ``spammers'', one can construct reasonable objective functions for crowdsourcing that guarantee convex inference.
\end{quote}
\end{abstract}

\section{Introduction}
Human computation (or crowdsourcing) involves humans performing tasks which are generally difficult for computers to perform. Since humans may not have perfect abilities and sometimes may not even have good intentions, machine learning and statistical inference algorithms are typically employed
to post-process the data obtained from the human workers in order to infer the true answers~\cite{dawid1979maximum,whitehill2009whose,welinder2010multidimensional,RayYuZha10,karger2011budget,wauthier2011bayesian,mccreadie2011crowdsourcing,luon2012rankr,zhou2012learning,liu2012variational,shah2013case,bachrach2012grade,kamar2012combining,vempaty2013icassp,salek2013hotspotting,shah2014better,matsui2013crowdsourcing,piech2013tuned,chen2013pairwise}.
These algorithms infer the true solutions to the tasks and the worker abilities by minimizing some carefully designed function that captures these parameters and their dependence on the observed responses. We shall call this function the ``\fname function''. For example, it is typical to assume a generative model and take a maximum likelihood estimation. In such a case, the objective function is the negative-likelihood or the negative-log-likelihood of the data obtained from the workers, and Expectation-Maximization (EM) style optimization procedures are often employed to minimize this function ~\cite{dawid1979maximum,whitehill2009whose,welinder2010multidimensional,RayYuZha10,liu2012variational,zhou2012learning,chen2013pairwise}.

\begin{figure}[t]
\centering
\includegraphics[width=.8\columnwidth]{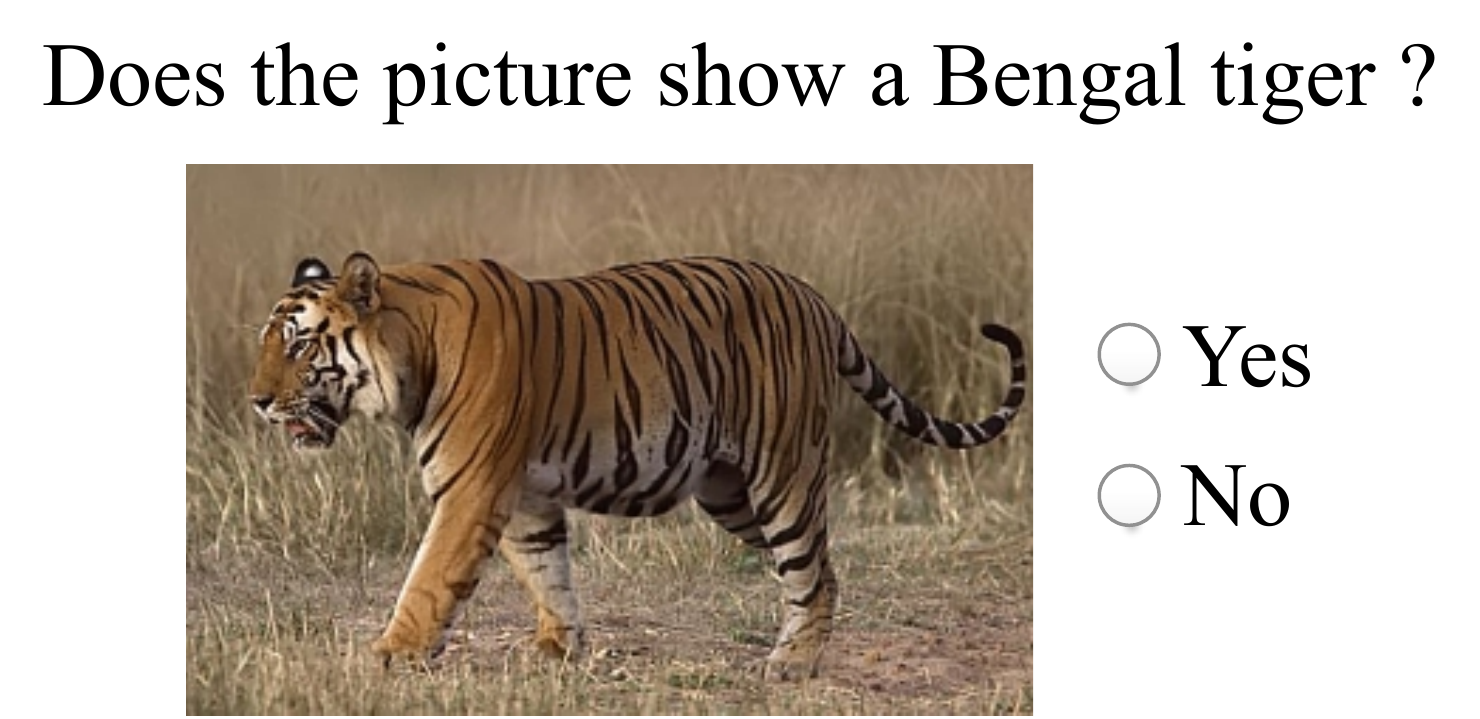}
\caption{An example of a binary-choice task.}
\label{fig:interfaces_binary}
\end{figure}


To the best our knowledge, none of the objective functions in the literature for the inference problem in human computation is convex, and there are no generic guarantees available regarding obtaining a global optimum. Inspired by the extremely successful machine learning algorithms which are developed with convex objective functions (such as support vector machines (SVMs)~\cite{vapnik1998statistical}),   
we investigate if there exists a meaningful convex objective function for the inference problem in human computation. Convex functions have appealing properties, and convex problems have been studied extensively in the literature, with high performance algorithms and rigorous guarantees available for very generic settings. It is thus of interest to investigate `reasonable' models for human computation that ensure convexity in inference, thereby providing the ability to take advantage of this vast body of literature. 

In this paper, we investigate the problem of convexity in human computation via an axiomatic approach. Our results show that unfortunately, under two mild and natural assumptions in crowdsourcing, no model can guarantee convexity in the inference procedure. We show subsequently that all known models for crowdsourcing satisfy the two proposed mild axioms. The takeaway from this result is that it is futile to construct human-computation models, for present setups, that attempt to gain tractability by ensuring convexity. 
Finally, we show that interestingly, if one can forgo the explicit modelling of ``spammers'', it is indeed possible to construct `reasonable' models for human computation that guarantee convexity in the objective for inference. 

\section{Problem Setting}
Consider a binary-choice setting, where the worker must select from two given options for every question (e.g., Figure~\ref{fig:interfaces_binary}).\footnote{Results on `impossibility of convexity' for the binary-choice setting extend to more general settings (such as multiple-choice with more than two choices) by restricting attention to only two choices.} Suppose there are $\numworkers$ workers, each of whom is assumed to have some latent `ability'. Denote the latent ability of worker $i \in [\numworkers]$ as $\work_i \in \workinterval$ for some $\lowerw<\upperw \in\mathbb{R}$, with a higher value of $\work_i$ representing a more able worker.\footnote{We adopt the standard notation of representing the set $\{1,2,\ldots,\alpha\}$ as $[\alpha]$ for any positive integer $\alpha$, and representing the interval of the real line between (and including) $\beta_1$ and $\beta_2$ as $[\beta_1,\beta_2]$ for any two real numbers $\beta_1<\beta_2$.} We will use the value ``$\spammerw$'' to represent the ability of a \emph{spammer} and assume that $\spammerw \in [\lowerw,\upperw)$. A spammer is a worker who answers randomly with no regard to the question being asked. Spammers are known to exist in plentiful numbers among the worker pools on crowdsourcing platforms~\cite{bohannon2011social,kazai2011crowdsourcing,vuurens2011much,wais2010towards}. Throughout the paper, we will restrict our attention to the (convex) subset $[0,\upperw]$ of the parameter space of the worker ability, since a function that is not convex on this set $[0,\upperw]$ will not be convex on the superset $\workinterval$ either. For convenience of notation, we define $\workbold := [\work_1 \cdots \work_\numworkers]^T \in [0,\upperw]^\numworkers$. 

There are $\numitems$ questions, and each question has two choices, say, ``$0$'' and ``$1$''. This paper looks at procedures to infer, \emph{via convex optimization}, the true answers $\quesbold^* \in \{0,1\}^\numitems$ to these $\numitems$ questions and the worker abilities $\workbold \in [0,\upperw]^\numworkers$ from the responses received from the workers. However, an optimization problem over a discrete set is non-convex by definition. A convex optimization based procedure will thus relax the discrete domain of $\mathbf{\ques}^*$ to a continuous set. To this end, we associate every question $j \in [\numitems]$ to a parameter $\ques_j \in [0,1]$. The inference procedure operates on $\quesbold := [\ques_1 \cdots \ques_\numitems]^T \in [0,1]^\numitems$, and the inferred values may be subsequently quantized to obtain a solution in the discrete set $\{0,1\}^\numitems$.  


Every question is asked to one or more of the $\numworkers$ workers, and every worker is asked one or more of the $\numitems$ questions. 
We shall represent the workers' responses as a $\{0,1,\infty\}$-valued matrix $\respmx$ of size $(\numworkers \times \numitems)$. For every $i \in [\numworkers]$ and $j\in [\numitems]$, we let $\respmx_{i,j}$ denote the $(i,j)^{\rm th}$ element of the matrix $\respmx$. The value of $\respmx_{i,j}$ is set as $\infty$ if worker $i$ was not asked question $j$, and is set as worker $i$'s response to question $j$ otherwise. The answers to the questions and the abilities of the workers are now inferred by minimizing some reasonable \fname function $\cost$ of $\quesbold$ and $\workbold$ given the worker-responses $\respmx$:
\[
(\hat{\quesbold},\hat{\workbold}) = \argmin_{\quesbold \in [0,1]^\numitems,\workbold \in [0,\upperw]^\numworkers} \cost\left(\quesbold,\workbold;\respmx\right).
\]
In the sequel, we will discuss the possible convexity of this optimization program under certain proposed axioms, and show that this optimization cannot be convex even after relaxing the domain of the answers from $\{0,1\}^\numitems$ to $[0,1]^\numitems$. 

Upon completion of the inference procedure, an additional rounding step is often executed to convert $\hat{\quesbold} \in [0,1]^\numitems$ to a discrete space $\hat{\quesbold}^* \in \{0,1\}^\numitems$. This rounding may be performed via a deterministic approach, for instance, by quantizing $\ques_j$ to $1$ iff $\hat{\ques}_j > 0.5$, or via a probabilistic approach, for instance, by quantizing $\hat{\ques}_j$ to $1$ with a probability $\hat{\ques}_j$, for every question $j \in [\numitems]$. Many a times, however, the inferred vector $\hat{\quesbold}$ is often left in the continuous set $[0,1]^\numitems$ as a ``soft'' output. In either case, the value $|\hat{\ques}_j-0.5|$ is interpreted as the ``confidence'' associated with the inference of the answer to the $j^{\rm th}$ question, with a higher value of $|\hat{\ques}_j-0.5|$ implying a higher confidence in the answer inferred for question $j$. \emph{A larger value of $\hat{\ques}_j$ indicates a greater confidence that $\ques^*_j$ is $1$. A higher value of $\hat{\work}_i$ for any $i \in [\numworkers]$ is interpreted as a greater belief in the ability of worker $i$.}

\section{Axiomatization of Objective Function}\label{sec:axioms}

We take an axiomatic approach towards the design of the \fname function $\cost$, and formulate two weak and natural axioms that the \fname function $\cost$ must satisfy. Recall that the inference procedure \textit{minimizes} the objective function $\cost$, and hence the objective function should have a \textit{lower} value when its arguments form a \textit{better fit} to the observed data. While the axioms are stated in a general manner below, for an intuitive understanding one may think of a maximum likelihood approach with $\cost(\quesbold,\workbold;\respmx)$ as the negative log likelihood of the observed data $\respmx$ conditioned on $(\quesbold,\workbold)$.

First consider the case when there is only $\numitems=1$ question and $\numworkers=1$ worker. The vectors $\quesbold$ and $\workbold$ can be represented by scalars $\ques \in [0,1]$ and $\work \in [0,\upperw]$ respectively, and the matrix $\respmx$ can be represented as a scalar in $\{0,1\}$.

\begin{axiom}[Distinguishing different worker abilities] \label{axiom:ball}
There exists $\epsilon>0$ such that $\cost(\ques,\work;1)>\cost(\ques,0;1)~\forall(\ques,\work)\in(0,\epsilon)\times (0,\epsilon)$ and $\cost(\ques,\work;1)<\cost(\ques,0;1)~\forall(\ques,\work)\in (1-\epsilon,1) \times (0,\epsilon)$.
\end{axiom}

Informally, Axiom~\ref{axiom:ball} says that if the worker reports `$1$' when $\ques$ is also close to $1$, then the worker is likely to be more able. On the other hand, if the worker reports the opposite answer, then the worker is likely to be less able.

\begin{axiom}[Modeling spammers]\label{axiom:spammer}
The \fname function $\cost(\ques,\work;1)$ is independent of $\ques$ when $\work=\spammerw$.
\end{axiom}
A spammer is a worker who answers randomly, without any regard to the question being posed. Spammers are highly abundant in today's crowdsourcing systems, and pose a major challenge to the data collection as well as the inference procedures. Axiom~\ref{axiom:spammer} necessitates an explicit incorporation of a spammer into the parameter space.


One may also define analogous axioms for the function $\cost(\cdot,\cdot;0)$, but for the purposes of this paper, it will suffice to work with solely the function $\cost(\cdot,\cdot;1)$. Observe that we have \textit{not} made any other assumptions on the function $\cost$ such as continuity, differentiability, or Lipchitzness.

In the next section, we will show that no \fname function $\cost$ for inference in human computation that satisfies the two simple requirements identified above can be convex. We will also show that all existing inference techniques for crowdsourcing (that we are aware of) fall into this class. We will also subsequently demonstrate, by means of a constructive example, the interesting fact that in the absence of the requirement of modelling a spammer, the \fname function can indeed be convex.

The three axioms listed above for the setting of $\numitems = \numworkers=1$ are translated to the general setting of $\numitems \geq 1$ questions and $\numworkers \geq 1$ workers in the following manner. Consider the convex subset of the domain of parameters $\quesbold$ and $\workbold$ that is given by $\quesbold = \ques \mathbf{1}$ and $\workbold = \work \mathbf{1}$, where $\ques \in [0,1], \work \in [0,\upperw]$. Furthermore, suppose the observed data is $Y_{i,j}=1$ for every $(i,j)$ for which worker $i$ is asked question $j$. In this case the \fname function reduces to being a function of only the scalars $\ques$ and $\work$, as $\cost(\ques,\work;1)$, and we can now call upon the two axioms listed above. Now, if an optimization problem is non-convex over a convex subset of the parameter space, then it is non-convex over the entire parameter space as well.  Thus, it suffices to show non-convexity for the case of $\numitems = \numworkers = 1$ and the result for $\numitems \geq 1,~\numworkers \geq 1$ follows.

\section{Impossibility of Convexity}

Theorem~\ref{thm:noconvex} below proves the impossibility of models guaranteeing convex inference in crowdsourcing.
\begin{theorem}\label{thm:noconvex}
No function $\cost$ satisfying Axiom~\ref{axiom:ball} and Axiom~\ref{axiom:spammer} can be convex.
\end{theorem}
Non-convex problems are often converted to convex problems via transformations of the variables involved. Our result however says that for any such transformation of the variables at hand, as long as the semantic meanings of the variables are retained, it is reasonable to expect the two aforementioned axioms to be satisfied, rendering the result of Theorem~\ref{thm:noconvex} applicable.

\section{Some Examples of Existing Models}
To the best of our knowledge, all existing models for inference in crowdsourcing tasks satisfy our two axioms. 
We illustrate the same with a few examples in this section.

While the two axioms of Section~\ref{sec:axioms} were constructed to identify the precise cause of non-convexity, in this section, we propose three general properties that we would like any \fname function for crowdsourcing to satisfy. As we will see later, satisfying these three properties automatically implies adherence to the two axioms. We will also show that all existing models satisfy these three properties. As before, while the properties are defined for generic human-computation models, for an easier understanding one may think of the \fname function $\cost$ as the negative log likelihood function.

\begin{property}[Monotonicity in accuracy of the answer]\label{prop:answer}
$\cost(\ques,\work;1)$ is non-increasing in $\ques$.
\end{property}
Property P\ref{prop:answer} says that the likelihood does not decrease if $\ques$ is brought closer to the observed response ``$1$ '' (recall that we are considering only the interval $[0,\upperw]$ for $\work$, where the latent ability of the worker is no worse than randomly answering).
\begin{property}[Monotonicity in worker ability] \label{prop:worker}
There exists some non-decreasing function $g:[0,\upperw) \rightarrow [0.5,1)$ such that $\cost(\ques,\work;1)$ increases with $\work$ when $\ques < g(\work)$ and $\cost(\ques,\work;1)$ decreases with an increase in $\work$ when $\ques > g(\work)$.
\end{property}
Recall that for an inferred answer $\hat{\ques}$, the value $|\hat{\ques}-0.5|$ represents the confidence associated to the inference. A higher confidence in the inference directly relates to a greater belief in the ability of a worker. Property~P\ref{prop:worker} formalizes this relation, with function $g(\work)$ capturing the confidence associated to the work of a worker with ability $\work$. The likelihood is thus higher when the confidence associated to the work of the worker is closer to the corresponding confidence in the inferred answer.
\begin{property}[Modeling spammers]\label{prop:spammer}
The \fname function $\cost(\ques,\work;1)$ is independent of $\ques$ when $\work=\spammerw$.
\end{property}

Property~P\ref{prop:spammer} is identical to Axiom~\ref{axiom:spammer}.
\begin{proposition}\label{proposition:properties_imply_axioms}
Any objective function $\cost$ that satisfies Property P\ref{prop:worker} also satisfies Axiom~\ref{axiom:ball}.
\end{proposition}

We now present examples of existing models for crowdsourcing and show that these models indeed satisfy the three properties listed above (and therefore the two axioms defined in Section~\ref{sec:axioms}). Throughout this section, we will let $\numobs$ denote the number of responses received from the workers, i.e, the number of $\{0,1\}$-valued entries in the response matrix $\respmx$.

\begin{example}[Dawid-Skene model]~\\
\noindent Model: The Dawid-Skene model is one of the most popular models for crowdsourcing~\cite{dawid1979maximum,ipeirotis2010quality,gao2013minimax,zhang2014spectral,karger2011budget,dalvi2013aggregating,ghosh2011moderates}. The model assumes that the ability of a worker represents the probability of her correctly answering any individual question, i.e., if worker $i \in [\numworkers]$ is asked question $j \in [\numitems]$, then she will give the correct answer with a probability $p_i$ and an incorrect answer with probability $(1-p_i)$, for some parameter $p_i \in [0,1]$ whose value is unknown. The response of any worker to any question is independent of all else. In order to ensure that the model is identifiable, it is typically also assumed that $p_i \in [0.5,1]~\forall i \in [\numworkers]$, or $\frac{1}{\numworkers} \sum_{i=1}^{\numworkers}p_i \geq 0.5$, or simply $p_1 \geq 0.5$. We shall restrict our attention to $p_i \in [0.5, 1]~\forall~i~\in [\numworkers]$. Further, we will work with a shifted and scaled version of $p_i$'s by defining $\work_i = 2p_i-1~\forall~i \in [\numworkers]$. Under this transformation, we have $\lowerw=0$ and $\upperw=1$.

Inference: Consider inferring $\quesbold$ and $\workbold$ via maximum likelihood estimation. Observe that the likelihood of observation $\respmx$ is
\begin{align*}
\prod_{(i,j):\respmx_{i,j}\neq \infty} &\left(\frac{1+\work_i}{2}\right)^{\respmx_{i,j} \ques_j + (1-\respmx_{i,j})(1-\ques_j)}\\ 
\times &\left(\frac{1-\work_i}{2}\right)^{\respmx_{i,j} (1-\ques_j) + (1-\respmx_{i,j})\ques_j}.
\end{align*}
{
\thinmuskip = .1\thinmuskip
\medmuskip = .1\medmuskip
\thickmuskip = .1\thickmuskip
The negative log-likelihood is thus given by
\begin{align*}
\cost(\quesbold,\workbold;\respmx) := -\hspace{-.3cm}\sum_{\substack{(i,j):\\Y_{i,j}\neq \infty}}\hspace{-.3cm} \left(\hspace{-.12cm}\left(\hspace{-.03cm}\respmx_{i,j} \ques_j + (1-\respmx_{i,j})(1-\ques_j)\right) \log \left(\!\!\!\!\frac{1+\work_i}{2}\!\!\!\!\right) \right.\\ 
\hspace{-4.6cm} \left.+ \left(\respmx_{i,j} (1-\ques_j) + (1-\respmx_{i,j})\ques_j\right) \log \left(\frac{1-\work_i}{2}\right)\right),
\end{align*}
and this function is minimized over $(\quesbold,\workbold) \in [0,1]^{\numitems+\numworkers}$.

Consider the following subspace of the arguments: $\quesbold = \ques \mathbf{1}$ and $\workbold = \work \mathbf{1}$ for some $\ques \in [0,1]$ and $\work \in [0,1]$. Further, suppose $\respmx_{i,j} =1$ for every $(i,j)$ for which worker $i$ was asked question $j$. Under this restriction, the \fname function reduces to
\begin{multline}
%
\hspace{-.4cm}\cost(\ques,\work; 1)= -\numobs \left(\ques \log \left(\frac{1+\work}{2}\right) + (1-\ques) \log \left(\frac{1-\work}{2}\right)\right)\hspace{-.25cm}\label{eq:dawid_skene_objective}
\end{multline}
}

Properties: Let us understand what the three properties listed earlier mean in this context. Observe that 
\begin{align*} \cost(\ques, 0; 1) &= -\numobs (\ques \log 0.5 + (1-\ques) \log 0.5) \\&= -\numobs \log 0.5, \end{align*} and hence $\cost(\ques,\work;1)$ is independent of $\ques$ when $\work=0$. Thus $\work=0$ models a spammer, and the function $\cost$ obeys Property P\ref{prop:spammer}. For Property P\ref{prop:answer}, observe that when $\work \in [0,1]$, 
\begin{align*} \frac{\partial}{\partial \ques}\cost(\ques, \work; 1) &= -\numobs \left(\log \left(\frac{1+\work}{2}\right) - \log \left(\frac{1-\work}{2}\right)\right)\\ & \leq 0.
\end{align*} 
Thus Property P\ref{prop:answer} is satisfied. Finally, 
\begin{align*}\frac{\partial}{\partial \work}\cost(\ques, \work; 1) = -\numobs \left( \frac{\ques}{1+\work} - \frac{1-\ques}{1-\work} \right),\end{align*} 
which is positive when $2\ques-1 < \work$ and negative when $2\ques-1 > \work$. Property P\ref{prop:worker} is thus satisfied with $\partition(\work) = \frac{1+\work}{2}$.

The \fname function~\eqref{eq:dawid_skene_objective} is plotted in Figure~\ref{fig:objective_dawidskene}. 

\begin{figure}
\centering
\includegraphics[width=.42\textwidth]{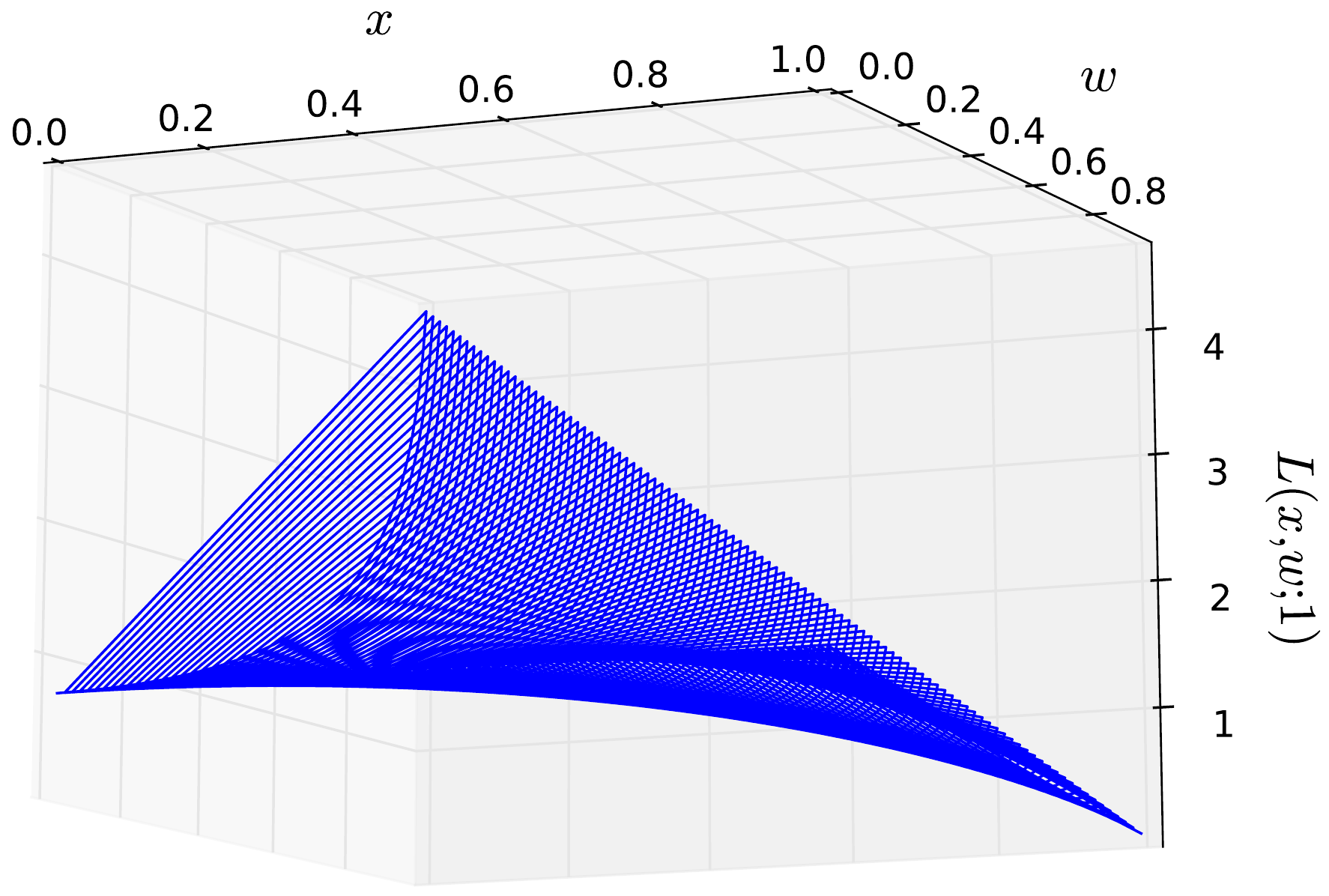}
\caption{The objective function for maximum-likelihood inference under the Dawid-Skene model.}
\label{fig:objective_dawidskene}
\end{figure}
\end{example}

\begin{example}[Two-coin Dawid-Skene model]~\\
\noindent The two-coin model associates the ability of every worker $i \in [\numworkers]$ to two latent variables $p_{i,0}\in [0.5,1]$ and $p_{i,1} \in [0.5,1]$. Under the two-coin model, the probability with which a worker answers a question correctly depends on the true answer to that question: if the true answer to a question is $\ques^* \in \{0,1\}$ then the worker (correctly) provides $\ques^*$ as the answer with probability $p_{i,\ques^*}$ and (incorrectly) provides $(1-\ques^*)$ as the answer otherwise, independent of all else. In order to connect to our theory, we simply restrict our attention to a (convex) subset of the parameters $\{p_{i,0},p_{i,1}\}_{i=1}^{\numworkers}$ obtained by setting $p_{i,0}=p_{i,1}~\forall~i\in[\numworkers]$. The resulting model is identical to the one-coin model discussed earlier.
\end{example}

\begin{example}[Additive Noise]~\\
Model: The additive noise model assumes that when worker $i \in [\numworkers]$ is asked question $j \in [\numitems]$, the response  $\respmx_{i,j}$ of the worker is given by 
\begin{align*}
\respmx_{i,j} = \mathds{1}\{\work_i (\ques_j-0.5) + \epsilon_{i,j}>0\},
\end{align*}
where $\{\epsilon_{i,j}\}_{i \in [\numitems], j \in [\numworkers]}$ is a set of i.i.d. random variables with some (known) c.d.f. $F$. The function $F$ is non-constant in the domain of interest $[-0.5\upperw,0.5\upperw]$. The response is assumed to be independent of all other questions and all other workers. A common choice for $F$ is the c.d.f. of the Gaussian distribution~\cite{piech2013tuned,thurstone1927law,welinder2010multidimensional}.

Inference: The inference is usually performed by minimizing the negative log likelihood of the observed data $\respmx$:
\begin{align*}
\argmin_{\quesbold,\workbold} ~~~ - \hspace{-.3cm} \sum_{(i,j):\respmx_{i,j} \neq \infty}\hspace{-.3cm} \left( \respmx_{i,j} \log (1-F(-(\ques_j-0.5) \work_i)) \right. \\ ~\qquad \qquad \left. +  (1-\respmx_{i,j}) \log (F(-(\ques_j-0.5) \work_i)\right).
\end{align*}

Properties: Let us now relate this model to the three properties enumerated earlier. Let us restrict our attention to the (convex) subset of the parameters where $\quesbold = \ques \mathbf{1}$ and $\workbold = \work \mathbf{1}$ for some $\ques \in [0,1]$ and $\work \in [\spammerw,\upperw]$. Suppose $\respmx_{i,j}=1$ for every $(i,j)$ for which worker $i$ was asked question $j$. The \fname function then reduces to
\begin{align*}
\cost(\ques,\work; 1) = - \numobs \log (1-F(-(\ques-0.5) \work)) .
\end{align*}
One can verify that this function is non-increasing in $\ques$ (whenever $\work \geq 0$), thereby satisfying Property P\ref{prop:answer}. Furthermore, setting $g(\work) = 0.5$ satisfies Property P\ref{prop:worker}. A spammer is modeled by the parameter value $\work=0$, in which case, the function $\cost$ ceases to be dependent on $\ques$.
\end{example}

\begin{example}[Minimax Entropy Model]~\\
Model: The minimax entropy model~\cite{zhou2012learning} hypothesizes that when worker $i \in [\numworkers]$ answers question $j \in [\numitems]$, she provides ``1'' as the answer with a probability $\pi_{i,j}$ and ``0'' otherwise, independent of all else, for some unknown value $\pi_{i,j} \in [0,1]$. Under the `minimax entropy principle' proposed therein, the set $\{\pi_{i,j}\}_{i \in [\numworkers], j\in [\numitems]}$ has the maximum entropy under the constraints imposed by the set of true answers and the observed data, and the true answers minimize this value of maximum entropy:
\begin{align*}
\min_{\quesbold} &\max_{\{\pi_{i,j}\}_{i \in [\numworkers], j\in [\numitems]}} -\sum_{i=1}^{\numworkers} \sum_{j=1}^{\numitems} \pi_{i,j} \ln \pi_{i,j}\\
{\rm s.t.~}&\sum_{i=1}^{\numworkers} \pi_{i,j} = \sum_{i=1}^{\numworkers} \respmx_{i,j}~\forall j \in [\numworkers]\\
&\sum_{j=1}^{\numitems} \ques_j \pi_{i,j} = \sum_{j=1}^{\numitems} \ques_j \respmx_{i,j}~\forall i \in [\numworkers]\\
&\sum_{j=1}^{\numitems} (1-\ques_j) \pi_{i,j} = \sum_{j=1}^{\numitems} (1-\ques_j) \respmx_{i,j}~\forall i \in [\numworkers]\\
&0 \leq \pi_{i,j}, \ques_j \leq 1 ~ \forall i \in [\numworkers],~j \in [\numitems].
\end{align*}

Inference: The authors show that the values $\pi_{i,j}$ must necessarily be of the form 
{
\thinmuskip = .5\thinmuskip
\medmuskip = .5\medmuskip
\thickmuskip = .5\thickmuskip
\begin{align*}
\pi_{i,j} = \frac{\exp\left( (1-\ques_j)(\tau_{j,1} + \sigma_{i,1,0}) + \ques_j (\tau_{j,1} + \sigma_{i,1,1})\right)}{\sum_{\ell=0}^{1}\exp\left( (1-\ques_j)(\tau_{j,\ell} + \sigma_{i,\ell,0}) + \ques_j (\tau_{j,\ell} + \sigma_{i,\ell,1})\right)}
\end{align*}
}
for some parameters $\{\sigma_{i,\ell_1,\ell_2}\}_{i \in [\numworkers],\ell_1 \in \{0,1\}, \ell_2 \in \{0,1\}}$ and $\{\tau_{j,\ell}\}_{j\in[\numitems],\ell \in \{0,1\}}$.
The authors then propose minimization (with respect to variables $\{\ques_i,\sigma_{i,\ell_1,\ell_2},\tau_{j,\ell}\}$) of the dual of the aforementioned program, which they derive to be of the form
{
\thinmuskip = .1\thinmuskip
\medmuskip = .1\medmuskip
\thickmuskip = .1\thickmuskip
\begin{align*}
-\hspace{-.25cm} \sum_{\substack{(i,j):\\\respmx_{i,j} \neq \infty}} \hspace{-.25cm} \frac{\exp\left( (1-\ques_j)(\tau_{j,\respmx_{i,j}} + \sigma_{i,\respmx_{i,j},0}) + \ques_j (\tau_{j,\respmx_{i,j}} + \sigma_{i,\respmx_{i,j},1})\right)}{\sum_{\ell=0}^{1}\exp\left( (1-\ques_j)(\tau_{j,\ell} + \sigma_{i,\ell,0}) + \ques_j (\tau_{j,\ell} + \sigma_{i,\ell,1})\right)}
\end{align*}
}

Properties: Consider the following (convex) subset of the parameter space: $\tau_{j,\ell}=0~\forall j \in [\numitems],\ell \in \{0,1\}$, $\sigma_{i,0,0}=\sigma_{i,1,1}=1-\sigma_{i,1,0}=1-\sigma_{i,0,1}:=0.5 + \work_i \in [0,1]~\forall i \in [\numworkers]$. The minimization program can now be rewritten as
\begin{align*}
\argmin_{\quesbold,\workbold \in [0,1]^{\numitems+\numworkers}} \sum \cost(\ques_i,\work_j;\respmx_{i,j})
\end{align*}
where
\begin{align*}
\cost(\ques,&\work;1) := \\
&\frac{e^{(1-\ques)(0.5-\work) + \ques (0.5+\work)}}{e^{(1-\ques)(0.5-\work) + \ques(0.5+\work)} + e^{\ques(0.5-\work) + (1-\ques) (0.5+\work)}},
\end{align*}
and $\cost(\ques,\work;0) = \cost(1-\ques,\work;1)$. 
With some algebraic manipulations, one can verify that this function $\cost(\cdot,\cdot;1)$ satisfies Property P\ref{prop:answer}, Property P\ref{prop:worker} (with $g(\work) = 0.5$) and Property P\ref{prop:spammer} (with $\work=0$ representing a spammer).
\end{example}

\begin{example}[GLAD model]~\\
Model and inference: The GLAD model was introduced in~\cite{whitehill2009whose}. We will restrict attention to the subspace of the parameter set which has, in the notation of~\cite{whitehill2009whose}, $\beta_j=0~\forall~j$. Using the notation of the present paper for the rest of the parameters, the objective function (the negative log likelihood) is
\begin{align*}
\cost(\ques,\work; 1) = \numobs \ques \log \left(1+ e^{-\work} \right) + \numobs (1-\ques) \log \left(1+ e^{\work} \right)
\end{align*}
under the convex subset $\quesbold = \ques \mathbf{1}$ and $\workbold = \work \mathbf{1}$ of the parameters, wth $\ques \in [0,1],\,\work \in [0,1]$. 

Properties: The derivative of $\cost(\ques,\work;1)$ with respect to $\ques$ is non-positive (for $\work \geq 0$), thus satisfying Property P\ref{prop:answer}. One can also verify that Property P\ref{prop:worker} is satisfied with $g(\work)=\frac{1}{1+e^{-\work}}$. Finally, we have $\cost(\ques,0;1)=\numobs \log 2$, thereby satisfying Property P\ref{prop:spammer}.
\end{example}

\begin{figure}
\centering
\includegraphics[width=.42\textwidth]{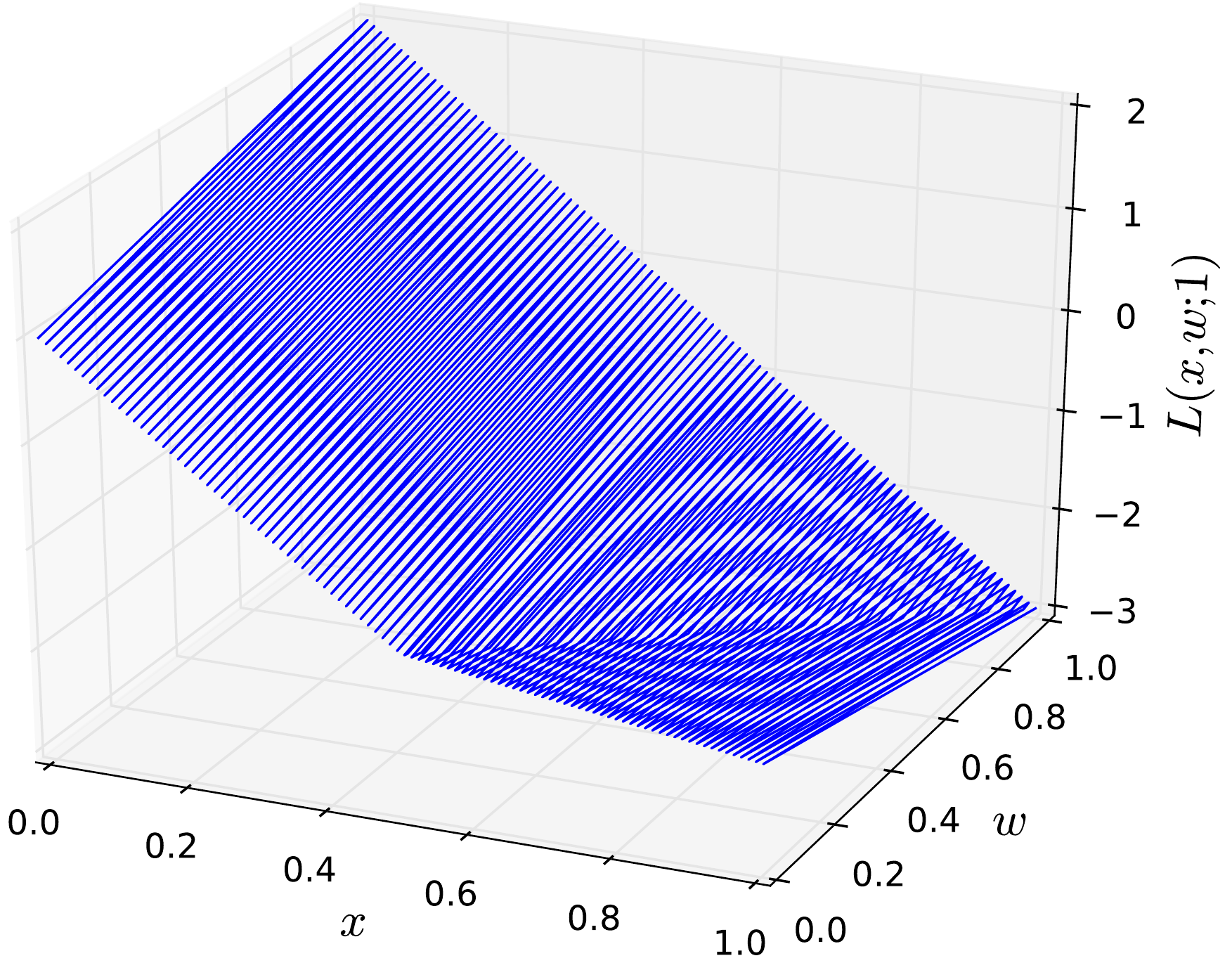}
\caption{An objective function that satisfies properties P\ref{prop:answer} and P\ref{prop:worker}, is convex, but does not incorporate the modeling of a spammer. This function is defined in~\eqref{eq:convexNoAxiom3_L}.}
\label{fig:convexNoAxiom3}
\end{figure}

\section{If Not Modelling Spammers}

We now discuss the role of modeling a spammer, i.e., Axiom~\ref{axiom:spammer}. We show that ignoring this axiom indeed allows for convex \fname functions for crowdsourcing, not only satisfying Axiom~\ref{axiom:ball}, but also satisfying the stronger properties P\ref{prop:answer} and P\ref{prop:worker}.
\begin{theorem}\label{thm:convexNoAxiom3}
Let $\lowerw=0$ and $\upperw=1$. 
The function $\cost:[0,1] \times [0,1] \rightarrow \mathbb{R}$ defined as
\begin{align}
\cost(\ques,\work;1) = 
\begin{cases}
- \work - \ques -1 & \textrm{if }\work \leq 2\ques-1\\
\work - 5\ques +1  & \textrm{if }\work \geq 2\ques-1
\end{cases}~
\label{eq:convexNoAxiom3_L}
\end{align}
satisfies properties P\ref{prop:answer} and P\ref{prop:worker} (and hence Axiom~\ref{axiom:ball}) and is (jointly) convex in its two arguments.
\end{theorem}

For the general setting of multiple workers and multiple questions, letting 
\[\cost(\ques,\work;0) = \cost(1-\ques,\work;1),\]
and
\[
\cost(\quesbold,\workbold;\respmx) = \sum_{(i,j):\respmx_{i,j} \neq \infty } \cost(\ques_i,\work_j; \respmx_{ij})~
\]
ensures that the function $\cost(\cdot,\cdot;\respmx)$ satisfies Axiom 1 and is convex in its arguments $(\quesbold,\workbold)$. The \fname function $\cost(\cdot,\cdot; 1)$ constructed in Theorem~\ref{thm:convexNoAxiom3} is plotted in Figure~\ref{fig:convexNoAxiom3}. 

Note that we do not intend to claim the proposed function~\eqref{eq:convexNoAxiom3_L} as a ``good'' \fname function to use for crowdsourcing. Instead, the takeaway from this section is that if one forgoes the inclusion of spammers in the \fname, then one may indeed be able to design a crowdsourcing model that is reasonable and permits convex inference.



\section{Discussion}\label{sec:discussion}
It is important to be aware of the limitations of the framework of this paper. Throughout the paper we assumed no prior knowledge or complexity controls on the parameter space. One may alternatively consider the inference problem in a Bayesian setting with non-uniform priors, or impose some convex regularization. In fact, if the regularizer is strictly convex, then giving it a sufficiently large weight can make the objective function convex, albeit perhaps at the expense of the model not capturing certain essential features of the problem. However, as long as the objective function continues to satisfy the two  axioms presented here, the conclusions drawn in this paper continue to apply.


%
While convexity is certainly desirable, absence of convexity certainly does not mean the complete absence of guarantees; indeed, there is a line of recent works~\cite{loh2013regularized,netrapalli2013phase,zhang2014spectral} which provide guarantees for non-convex problems as well. 
In particular, although existing models for human computation are not convex, there exist theoretical guarantees on inference under the popular Dawid-Skene model to a certain extent~\cite{ghosh2011moderates,karger2011budget,dalvi2013aggregating,gao2013minimax,zhang2014spectral}. For instance,~\cite{zhang2014spectral} show that the EM algorithm for the Dawid-Skene model can achieve a minimax rate up to a logarithmic factor when it is appropriately initialized by spectral methods. However, these results need certain conditions which may not hold in real scenarios. Moreover, algorithms that are minimax optimal may not always work well in practice, for instance, see the experimental results in~\cite{liu2012variational}. 
Most importantly, guarantees for non-convex problems are constructed to-date on a case-by-case basis (for example, all known theoretical guarantees for crowdsourcing are for the Dawid-Skene model alone). These guarantees do not allow for a convenient application of the theory to any new model. On the other hand, although the theory of convex optimization is highly generic and extensive, the results of this paper imply that unfortunately one cannot readily exploit this theory in the context of human computation. 

This paper also shows that a willingness to forgo the explicit incorporation of spammers into the crowdsourcing model indeed allows for reasonable objective functions guaranteeing convex inference. Successful deterrence of spammers in crowdsourcing systems, for instance, by designing suitable reward mechanisms, may thus expand the scope of model-design for human computation. In conclusion, we would like to enumerate, partially in jest, the problems resulting from spammers in crowdsourcing systems: (a) low-quality work, (b) depletion of the monetary budget, and now, (c) prevention of models guaranteeing convex inference.

\section*{Appendix: Proofs}
\noindent\textit{Proof of Theorem~\ref{thm:noconvex}:} 
The proof employs a contradiction-based argument. Suppose there exists some function $\cost(\cdot,\cdot;1)$ that satisfies the two axioms and is convex in its two arguments in the set $[0,1] \times [0,\upperw]$. Without loss of generality assume $\epsilon \in (0, \min\{0.5,\upperw\})$.  Axiom~\ref{axiom:ball} mandates 
\begin{align}
\cost(\epsilon-\epsilon^2/2, \epsilon^2/2;1)&>\cost(\epsilon-\epsilon^2/2,0;1) \qquad \mbox{and}\label{eq:noconvexproof_0}\\
\cost(1-\epsilon/2,\epsilon/2;1)&<\cost(1-\epsilon/2,0;1).\label{eq:noconvexproof_1}
\end{align} 
The assumed convexity of function $\cost$ implies
{
\thinmuskip = .1\thinmuskip
\medmuskip = .1\medmuskip
\thickmuskip = .1\thickmuskip
\begin{align*}
(1-\epsilon)\cost(0,0;1) + \epsilon \cost(1-\epsilon/2,\epsilon/2;1) & \geq \cost(\epsilon-\epsilon^2/2, \epsilon^2/2;1).
\end{align*}
}
Substituting~\eqref{eq:noconvexproof_0} and~\eqref{eq:noconvexproof_1} in this inequality gives
\begin{align*}
(1-\epsilon)\cost(0,0;1) + \epsilon \cost(1-\epsilon/2,0;1) &> \cost(\epsilon-\epsilon^2/2, 0;1).
\end{align*}
Now, calling upon Axiom~\ref{axiom:spammer} gives
\begin{align*}
(1-\epsilon)\cost(0,0;1) + \epsilon \cost(0,0;1) &>\cost(0,0;1),
\end{align*}
yielding the desired contradiction.
\qed

~\\
\noindent\textit{Proof of Proposition~\ref{proposition:properties_imply_axioms}:} Pick a value $\delta$ arbitrarily in $(0,\min\{1,\upperw\}/2)$. Set 
\begin{align*}
\epsilon = \min\{\delta, 1 - g(\delta)\}.
\end{align*} 
Observe that due to Property P\ref{prop:worker}, $\forall~\ques < 0.5$ (which includes all $\ques<\epsilon$), $\cost(\ques,\work;1)$ increases with an increase in $\work$, and hence $\cost(\ques,\work;1)>\cost(\ques,0;1)$ when $\work>0$. Also, for our chosen $\epsilon$, Property P\ref{prop:worker} implies that for any $\ques  \in (1 - \epsilon,1)$ and $\work < \epsilon$, the function $\cost(\ques,\work;1)$ decreases with an increase in $\work$. Thus we have $\cost(\ques,\work;1)<\cost(\ques,0;1)~\forall~(\ques,\work) \in (1-\epsilon,1)\times (0,\epsilon)$. 
Axiom~\ref{axiom:ball} is thus satisfied.
\qed

~\\
\noindent\textit{Proof of Theorem~\ref{thm:convexNoAxiom3}:}
Let us first verify that the proposed function $\cost(\cdot,\cdot; 1)$ satisfies the two properties. First, observe from the definition of $\cost$ in~\eqref{eq:convexNoAxiom3_L} that $\cost$ is always (strictly) decreasing in its first argument, and hence satisfies Property P\ref{prop:answer}. Towards Property P\ref{prop:worker}, set 
\begin{align*} g(\work) = (1+\work)/2.
\end{align*} 
By definition, $\cost(\ques,\work;1)$ (strictly) decreases with an increase in $\work$ when $\ques > g(\work)$, and (strictly) increases with an increase in $\work$ when $\ques < g(\work)$. 
Proposition~\ref{proposition:properties_imply_axioms} now guarantees that the function also satisfies Axiom~\ref{axiom:ball}.

Let us now investigate the convexity of this function. 
Consider two hyperplanes $H_0$ and $H_1$ defined as
\begin{align*}
H_0(\ques,\work) &= -\work - \ques - 1\\
H_1(\ques,\work) &= \work - 5\ques + 1.
\end{align*}
Observe that for any $(\ques,\work)$,
\begin{align*}
H_0(\ques,\work) - H_1(\ques,\work)  &= 2(-\work + 2\ques - 1).
\end{align*}
Thus, $H_0(\ques,\work) \geq H_1(\ques,\work)$ if $\ques \geq (1+\work)/2$ and $H_0(\ques,\work) \leq H_1(\ques,\work)$ if $\ques \leq (1+\work)/2$. It follows that 
\begin{align*}
\cost(\ques,\work;1) = \max\{H_0(\ques,\work), H_1(\ques,\work)\}
\end{align*}
meaning that $\cost$ is the maximum of two linear functions. Hence $\cost$ is convex.
\qed

\fontsize{9.5pt}{10.5pt} \selectfont

\bibliographystyle{aaai}

\begin{thebibliography}{}

\bibitem[\protect\citeauthoryear{Bachrach \bgroup et al\mbox.\egroup
  }{2012}]{bachrach2012grade}
Bachrach, Y.; Graepel, T.; Minka, T.; and Guiver, J.
\newblock 2012.
\newblock How to grade a test without knowing the answers---a {B}ayesian
  graphical model for adaptive crowdsourcing and aptitude testing.
\newblock In {\em ICML}.

\bibitem[\protect\citeauthoryear{Bohannon}{2011}]{bohannon2011social}
Bohannon, J.
\newblock 2011.
\newblock Social science for pennies.
\newblock {\em Science} 334(6054):307--307.

\bibitem[\protect\citeauthoryear{Chen \bgroup et al\mbox.\egroup
  }{2013}]{chen2013pairwise}
Chen, X.; Bennett, P.~N.; Collins-Thompson, K.; and Horvitz, E.
\newblock 2013.
\newblock Pairwise ranking aggregation in a crowdsourced setting.
\newblock In {\em ACM international conference on Web search and data mining},
  193--202.

\bibitem[\protect\citeauthoryear{Dalvi \bgroup et al\mbox.\egroup
  }{2013}]{dalvi2013aggregating}
Dalvi, N.; Dasgupta, A.; Kumar, R.; and Rastogi, V.
\newblock 2013.
\newblock Aggregating crowdsourced binary ratings.
\newblock In {\em International conference on World Wide Web},  285--294.

\bibitem[\protect\citeauthoryear{Dawid and Skene}{1979}]{dawid1979maximum}
Dawid, A.~P., and Skene, A.~M.
\newblock 1979.
\newblock Maximum likelihood estimation of observer error-rates using the {EM}
  algorithm.
\newblock {\em Applied statistics}  20--28.

\bibitem[\protect\citeauthoryear{Gao and Zhou}{2013}]{gao2013minimax}
Gao, C., and Zhou, D.
\newblock 2013.
\newblock Minimax optimal convergence rates for estimating ground truth from
  crowdsourced labels.
\newblock {\em arXiv preprint arXiv:1310.5764}.

\bibitem[\protect\citeauthoryear{Ghosh, Kale, and
  McAfee}{2011}]{ghosh2011moderates}
Ghosh, A.; Kale, S.; and McAfee, P.
\newblock 2011.
\newblock Who moderates the moderators?: {C}rowdsourcing abuse detection in
  user-generated content.
\newblock In {\em ACM conference on Electronic commerce},  167--176.
\newblock ACM.

\bibitem[\protect\citeauthoryear{Ipeirotis, Provost, and
  Wang}{2010}]{ipeirotis2010quality}
Ipeirotis, P.; Provost, F.; and Wang, J.
\newblock 2010.
\newblock Quality management on {A}mazon {M}echanical {T}urk.
\newblock In {\em ACM SIGKDD workshop on human computation},  64--67.

\bibitem[\protect\citeauthoryear{Kamar, Hacker, and
  Horvitz}{2012}]{kamar2012combining}
Kamar, E.; Hacker, S.; and Horvitz, E.
\newblock 2012.
\newblock Combining human and machine intelligence in large-scale
  crowdsourcing.
\newblock In {\em International Conference on Autonomous Agents and Multiagent
  Systems},  467--474.

\bibitem[\protect\citeauthoryear{Karger, Oh, and Shah}{2011}]{karger2011budget}
Karger, D.~R.; Oh, S.; and Shah, D.
\newblock 2011.
\newblock Budget-optimal crowdsourcing using low-rank matrix approximations.
\newblock In {\em Allerton Conference on Communication, Control, and
  Computing},  284--291.

\bibitem[\protect\citeauthoryear{Kazai \bgroup et al\mbox.\egroup
  }{2011}]{kazai2011crowdsourcing}
Kazai, G.; Kamps, J.; Koolen, M.; and Milic-Frayling, N.
\newblock 2011.
\newblock Crowdsourcing for book search evaluation: impact of {HIT} design on
  comparative system ranking.
\newblock In {\em ACM SIGIR conference on Research and development in
  Information Retrieval},  205--214.

\bibitem[\protect\citeauthoryear{Liu, Peng, and
  Ihler}{2012}]{liu2012variational}
Liu, Q.; Peng, J.; and Ihler, A.
\newblock 2012.
\newblock Variational inference for crowdsourcing.
\newblock In {\em NIPS},  701--709.

\bibitem[\protect\citeauthoryear{Loh and Wainwright}{2013}]{loh2013regularized}
Loh, P.-L., and Wainwright, M.~J.
\newblock 2013.
\newblock Regularized m-estimators with nonconvexity: Statistical and
  algorithmic theory for local optima.
\newblock In {\em NIPS},
  476--484.

\bibitem[\protect\citeauthoryear{Luon, Aperjis, and
  Huberman}{2012}]{luon2012rankr}
Luon, Y.; Aperjis, C.; and Huberman, B.~A.
\newblock 2012.
\newblock Rankr: A mobile system for crowdsourcing opinions.
\newblock In {\em Mobile Computing, Applications, and Services}. Springer.
\newblock  20--31.

\bibitem[\protect\citeauthoryear{Matsui \bgroup et al\mbox.\egroup
  }{2013}]{matsui2013crowdsourcing}
Matsui, T.; Baba, Y.; Kamishima, T.; and Kashima, H.
\newblock 2013.
\newblock Crowdsourcing quality control for item ordering tasks.
\newblock In {\em AAAI HCOMP}.

\bibitem[\protect\citeauthoryear{McCreadie, Macdonald, and
  Ounis}{2011}]{mccreadie2011crowdsourcing}
McCreadie, R.; Macdonald, C.; and Ounis, I.
\newblock 2011.
\newblock Crowdsourcing blog track top news judgments at TREC.
\newblock In {\em Workshop on crowdsourcing for search and data mining at the
  ACM international conference on web search and data mining (WSDM)},  23--26.

\bibitem[\protect\citeauthoryear{Netrapalli, Jain, and
  Sanghavi}{2013}]{netrapalli2013phase}
Netrapalli, P.; Jain, P.; and Sanghavi, S.
\newblock 2013.
\newblock Phase retrieval using alternating minimization.
\newblock In {\em NIPS},
  2796--2804.

\bibitem[\protect\citeauthoryear{Piech \bgroup et al\mbox.\egroup
  }{2013}]{piech2013tuned}
Piech, C.; Huang, J.; Chen, Z.; Do, C.; Ng, A.; and Koller, D.
\newblock 2013.
\newblock Tuned models of peer assessment in {MOOC}s.
\newblock {\em arXiv preprint arXiv:1307.2579}.

\bibitem[\protect\citeauthoryear{Raykar \bgroup et al\mbox.\egroup
  }{2010}]{RayYuZha10}
Raykar, V.~C.; Yu, S.; Zhao, L.~H.; Valadez, G.~H.; Florin, C.; Bogoni, L.; and
  Moy, L.
\newblock 2010.
\newblock Learning from crowds.
\newblock {\em Journal of Machine Learning Research} 11:1297--1322.

\bibitem[\protect\citeauthoryear{Salek, Bachrach, and
  Key}{2013}]{salek2013hotspotting}
Salek, M.; Bachrach, Y.; and Key, P.
\newblock 2013.
\newblock Hotspotting--a probabilistic graphical model for image object
  localization through crowdsourcing.
\newblock In {\em Twenty-Seventh AAAI Conference on Artificial Intelligence}.

\bibitem[\protect\citeauthoryear{Shah \bgroup et al\mbox.\egroup
  }{2013}]{shah2013case}
Shah, N.~B.; Bradley, J.~K.; Parekh, A.; Wainwright, M.; and Ramchandran, K.
\newblock 2013.
\newblock A case for ordinal peer-evaluation in {MOOC}s.
\newblock In {\em NIPS Workshop on Data Driven Education}.

\bibitem[\protect\citeauthoryear{Shah \bgroup et al\mbox.\egroup
  }{2014}]{shah2014better}
Shah, N.~B.; Balakrishnan, S.; Bradley, J.; Parekh, A.; Ramchandran, K.; and
  Wainwright, M.
\newblock 2014.
\newblock When is it better to compare than to score?
\newblock {\em arXiv preprint arXiv:1406.6618}.

\bibitem[\protect\citeauthoryear{Thurstone}{1927}]{thurstone1927law}
Thurstone, L.
\newblock 1927.
\newblock A law of comparative judgment.
\newblock {\em Psychological review} 34(4):273.

\bibitem[\protect\citeauthoryear{Vapnik}{1998}]{vapnik1998statistical}
Vapnik, V.~N.
\newblock 1998.
\newblock {\em Statistical learning theory}.
\newblock Wiley, NY.

\bibitem[\protect\citeauthoryear{Vempaty, Varshney, and
  Varshney}{2013}]{vempaty2013icassp}
Vempaty, A.; Varshney, L.~R.; and Varshney, P.~K.
\newblock 2013.
\newblock Reliable classification by unreliable crowds.
\newblock In {\em IEEE International Conference on Acoustics, Speech and Signal
  Processing (ICASSP)},  5558--5562.

\bibitem[\protect\citeauthoryear{Vuurens, de Vries, and
  Eickhoff}{2011}]{vuurens2011much}
Vuurens, J.; de~Vries, A.~P.; and Eickhoff, C.
\newblock 2011.
\newblock How much spam can you take? {A}n analysis of crowdsourcing results to
  increase accuracy.
\newblock In {\em ACM SIGIR Workshop on Crowdsourcing for Information
  Retrieval},  21--26.

\bibitem[\protect\citeauthoryear{Wais \bgroup et al\mbox.\egroup
  }{2010}]{wais2010towards}
Wais, P.; Lingamneni, S.; Cook, D.; Fennell, J.; Goldenberg, B.; Lubarov, D.;
  Marin, D.; and Simons, H.
\newblock 2010.
\newblock Towards building a high-quality workforce with {M}echanical {T}urk.
\newblock {\em NIPS workshop on computational social science and the wisdom of
  crowds}.

\bibitem[\protect\citeauthoryear{Wauthier and
  Jordan}{2011}]{wauthier2011bayesian}
Wauthier, F.~L., and Jordan, M.
\newblock 2011.
\newblock {B}ayesian bias mitigation for crowdsourcing.
\newblock In {\em NIPS},  1800--1808.

\bibitem[\protect\citeauthoryear{Welinder \bgroup et al\mbox.\egroup
  }{2010}]{welinder2010multidimensional}
Welinder, P.; Branson, S.; Perona, P.; and Belongie, S.~J.
\newblock 2010.
\newblock The multidimensional wisdom of crowds.
\newblock In {\em NIPS},
  2424--2432.

\bibitem[\protect\citeauthoryear{Whitehill \bgroup et al\mbox.\egroup
  }{2009}]{whitehill2009whose}
Whitehill, J.; Ruvolo, P.; Wu, T.-f.; Bergsma, J.; and Movellan, J.
\newblock 2009.
\newblock Whose vote should count more: Optimal integration of labels from
  labelers of unknown expertise.
\newblock In {\em NIPS},
  2035--2043.

\bibitem[\protect\citeauthoryear{Zhang \bgroup et al\mbox.\egroup
  }{2014}]{zhang2014spectral}
Zhang, Y.; Chen, X.; Zhou, D.; and Jordan, M.~I.
\newblock 2014.
\newblock Spectral methods meet {EM}: A provably optimal algorithm for
  crowdsourcing.
\newblock In {\em NIPS}.

\bibitem[\protect\citeauthoryear{Zhou \bgroup et al\mbox.\egroup
  }{2012}]{zhou2012learning}
Zhou, D.; Platt, J.; Basu, S.; and Mao, Y.
\newblock 2012.
\newblock Learning from the wisdom of crowds by minimax entropy.
\newblock In {\em NIPS},
  2204--2212.

\end{thebibliography}

\end{document}